\documentclass[12pt,reqno]{amsart}
\usepackage{fullpage}
\usepackage{amsmath}
\usepackage{mathrsfs,amssymb,graphicx,verbatim}
\usepackage{paralist}
\usepackage{amsfonts}
\usepackage{amssymb}
\usepackage{multirow}
\usepackage{euler, latexsym}

\usepackage{ifthen}

\newcommand{\ifsodaelse}[2]{\ifthenelse{\isundefined{\SODAF}}{#2}{#1}}

\ifsodaelse    {\usepackage{ltexpprt}\usepackage{amsmath}}{}
\usepackage{mathrsfs,amssymb,graphicx,verbatim}
\usepackage{paralist}
\usepackage[breaklinks,colorlinks,pdfstartview=FitH]{hyperref}

\newcommand\remove[1]{}

\newcommand{\rnote}[1]{}
\newcommand{\jnote}[1]{}

\date{}

\include{psfig}

\title[Influences of Standardization on Registration]{The Influence of Intensity Standardization on Medical Image Registration}

\author{Ula\c{s} Ba\u{g}c\i}
\address{School of Computer Science\\ University of Nottingham}
\email{ulasbagc@gmail.com}

\author{Jayaram~K.~Udupa}
\address{Medical Image Processing Group\\UPENN}
\email{jay@mipg.upenn.edu}
\author{ Li Bai}
\address{School of Computer Science\\University of Nottingham}
\email{bai@cs.nott.ac.uk}
\date{}

\begin{document}
\maketitle

\begin{abstract}
Acquisition-to-acquisition signal intensity variations (non-standardness) are inherent in MR images. Standardization is a post processing method for correcting inter-subject intensity variations through transforming all images from the given image gray scale into a standard gray scale wherein similar intensities achieve similar tissue meanings. The lack of a standard image intensity scale in MRI leads to many difficulties in tissue characterizability, image display, and analysis, including image segmentation. This phenomenon has been documented well; however, effects of standardization on medical image registration have not been studied yet. In this paper, we investigate the influence of intensity standardization in registration tasks with systematic and analytic evaluations involving clinical MR images. We conducted nearly 20,000 clinical MR image registration experiments and evaluated the quality of registrations both quantitatively and qualitatively. The evaluations show that intensity variations between images degrades the accuracy of  registration performance. The results imply that the accuracy of  image registration not only depends on spatial and geometric similarity but also on the similarity of the intensity values for the same tissues in different images.
\end{abstract}

\setcounter{tocdepth}{3}
\tableofcontents
\section{Introduction}
Image registration is an essential operation in a variety of medical imaging applications including disease diagnosis, longitudinal studies, data fusion, image segmentation, image guided therapy, volume reconstruction, pathology detection, and shape measurement~(\cite{nonmedsurvey,survey,survey2}). It is the process of finding a geometric transformation between a pair of scenes, the \textit{source scene} and the \textit{target scene}, such that the similarity between the transformed source scene (\textit{registered source}) and target scene becomes optimum. 
There are many challenges in the registration of medical images. Among these, those that stem from the artifacts associated with images include the presence of noise, interpolation artifacts, intensity non-uniformities, and intensity non-standardness. Although considerable research has gone into addressing the effects of noise~(\cite{holden}), interpolation~(\cite{reg_interp, reg_interp2, interp_likar}), and non-uniformity in image registration~(\cite{reg_nu}), little attention has been paid to study the effects of image intensity standardization/non-standardness in image registration. This aspect constitutes the primary focus of this paper.

MR image intensities do not possess a tissue specific numeric meaning even in images acquired for the same subject, on the same scanner, for the same body region, by using the same pulse sequence~(\cite{std, std_numeric}). Not only a registration algorithm needs to capture both large and small scale image deformations, but it also has to deal with global and local image intensity variations. The lack of a standard and quantifiable interpretation of image intensities may cause the geometric relationship between homologous points in MR images to be affected considerably. Current techniques to overcome these differences/variations fall into two categories. The first class of methods uses intensity modelling and/or attempts to capture intensity differences during the registration process. The second group constitutes post processing methods that are independent of registration algorithms. Notable studies that have attempted to solve this problem within the first class are~(\cite{demons,elastic,ash}). While global intensity differences are modelled with a linear multiplicative term in~(\cite{ash}), local intensity differences are modelled with basis functions. In~(\cite{elastic}), a locally affine but globally smooth transformation model has been developed in the presence of intensity variations which captures intensity variations with explicitly defined parameters. In~(\cite{demons}), intensities of one image are mapped into those of another via an adaptive transformation function. Although incorporating intensity modelling into the registration algorithms improves the accuracy, simultaneous estimation of intensity and geometric changes can be quite difficult and computationally expensive. 

The papers that belong to the second group of methods are~(\cite{std, std_numeric, std_new, std_nu, std_mtr}) in which a two-step method is devised for standardizing the intensity scale in such a way that for the same MRI protocol and body region, similar intensities achieve similar tissue meaning. The methods transform image intensities non-linearly so that the variation of the overall mean intensity of the MR images within the same tissue region across different studies obtained on the same or different scanners is minimized significantly. Furthermore, the computational cost of these methods is considerably small in comparison to methods belonging to the first class. Once tissue specific meanings are obtained, quantification and image analysis techniques, including registration, segmentation, and filtering, become more accurate. 

The non-standardness issue was first demonstrated in~(\cite{std}) where a method was proposed to overcome this problem. The new variants of this method are studied in~(\cite{std_new}). Numerical  tissue characterizability of different tissues is achieved by standardization and it is shown that this can significantly facilitate image segmentation and analysis in~(\cite{std_numeric, zhuge_seg}). Combined effects of non-uniformity correction and standardization are studied in~(\cite{std_nu}) and the sequence of operations to produce the best overall \textit{image quality} is studied via an interplaying sequence of non-uniformity correction and standardization methods. In~(\cite{std_gscale}), an improved standardization method based on the concept of generalized scale is presented. In~(\cite{std_mtr}), the performance of standardization methods is compared with the known tissue characterizing property of magnetization transfer ratio (MTR) imaging and it is demonstrated that tissue specific intensities may help characterizing diseases. 

The motivation for the research reported in this paper is the preliminary indication in~(\cite{bagci07}) of the potential impact that intensity standardization may have on registration accuracy. Currently no published study exists that has examined how intensity non-standardness alone may affect registration. The goal of this paper is, therefore, to study the effect of non-standardness on registration in isolation. Toward this goal, first intensity non-uniformities are corrected in a set of images, and subsequently, they are standardized to yield a ``clean set'' of images. Different levels of non-standardness are then introduced artificially into these images which are then subjected to known levels of affine deformations. The clean set is also subjected to the same deformations. The deformed images with and without non-standardness are separately registered to clean images and the differences in their registration accuracy are quantified to express the influence of non-standardness. The underlying methods are described in Section II and the analysis results are presented in Section III. Section IV presents some concluding remarks.
\section{Methods }
\label{sec:methods}
\subsection{Notations and Overview}
\label{subsec:term}
We represent a 3D image, called \textit{scene} for short, by a pair $\mathcal{C}=(C,f)$ where $C$ is a finite 3D array of voxels, called \textit{scene domain} of  $\mathcal{C}$, covering a body region of the particular patient for whom image data $\mathcal{C}$ are acquired, and $f$ is an intensity function defined on $C$, which assigns an integer intensity value to each voxel $\nu \in C$. We assume that $f(\nu) \geq 0$ for all $\nu \in C$ and $f(\nu)=0$ if and only if there are no measured data for voxel $\nu$. 

In dealing with standardization issues, the body region and imaging protocol need to be specified. All images that are analyzed for their dependence on non-standardness for registration accuracy are assumed to come from the same body region $B$ and acquired as per the same acquisition protocol $P$. The non-standardness phenomenon is predominant mainly in MR imaging. Hence, all image data sets considered in this paper pertain to MRI. However, the methods described here are applicable to any modality where this phenomenon occurs (such as radiography and electron microscopy).

There are six main components to the methods presented in this paper: (1) intensity non-uniformity correction, referred to simply as \textit{correction} and denoted by an operator $\kappa$; (2) intensity standardization denoted by an operator $ \psi $; (3) an affine transformation of the scene, denoted by $T$ used for the purpose of creating mis-registered scenes; (4) introduction of artificial intensity non-standardness denoted by the operator $\overline{\psi}$; (5) an affine scene transformation that is intended to register a scene with its mis-registered version; (6) evaluation methods used to quantify the goodness of scene registration.

Super scripts $c,s,\overline{s},t$ and $r$ are used to denote, respectively, the scenes resulting from applying correction, standardization, introduction of non-standardness, mis-registration, and registration operations to a given scene. Examples: 
$\mathcal{C}^c=\kappa\left( \mathcal{C}\right) ; \mathcal{C}^{cs}=\kappa\psi\left( \mathcal{C}\right); \mathcal{C}^{cs\overline{s}}=\overline{\psi}\left( \mathcal{C}^{cs}\right); \mathcal{C}^{cs\overline{s}t}=T\left( \mathcal{C}^{cs\overline{s}}\right)$. When a registration operation $T_r$ is applied to a scene $\mathcal{C}$, the target scene to which $\mathcal{C}$ is registered will be evident from the context. The same notations are extended to sets of scenes. For example, if $\chi$ is a given set of scenes for body region $B$ and protocol $P$, then $\chi^{cs\overline{s}}=\overline{\psi}\left(\chi^{cs} \right) $, where  $\chi^{cs}=\kappa\psi\left(\chi \right)$

Our approach to study the effect of non-standardness on registration is as follows:

\noindent (S1) Take a set $\chi$ of scenes, pertaining to a fixed $B$ and $P$, but acquired from different subjects in routine clinical settings.

\noindent (S2) Apply correction followed by standardization to the scenes in $\chi$ to produce the set $\chi^{cs}$ of \textit{clean scenes}. $\chi^{cs}$ is as free from non-uniformities, and more importantly, from non-standardness, as we can practically make. As justified in~(\cite{std_nu}), the best order and sequence of these operations to employ in terms of reducing non-uniformities and non-standardness is $\kappa$ followed by $\psi$. This is mainly because any correction operation introduces its own non-standardness.

\noindent (S3) Apply different known levels of non-standardness to the scenes in $\chi^{cs}$ to produce the set $\chi^{cs\overline{s}}$.

\noindent (S4) Apply different known levels of affine deformations $T$ to the scenes in $\chi^{cs\overline{s}}$ to form the scene set  $\chi^{cs\overline{s}t}$. Apply the same deformations to the clean scenes in the set $\chi^{cs}$ to create $\chi^{cst}$. In this manner for any scene $\mathcal{C}^{cs}\in\chi^{cs}$, we have the same scene after applying some non-standardness and the same deformation $T$, namely $\mathcal{C}^{cs\overline{s}t}$.

\noindent (S5) Register each scene $\mathcal{C}^{cs}\in\chi^{cs}$ to $\mathcal{C}^{cst}\in\chi^{cst}$ and determine the required affine deformation $T_s$ (the subscript s indicates ``standardized"). Similarly register each $\mathcal{C}^{cs}\in\chi^{cs}$ to  $\mathcal{C}^{cs\overline{s}t}\in\chi^{cs\overline{s}t}$ and determine affine deformation $T_{ns}$ (ns for ``not standardized") needed.

\noindent (S6) Analyze the deviations of $T_s$ and $T_{ns}$ from the true applied transformation $T$ over all scenes and as a function of the applied level of non-standardness and affine deformations.

In the rest of this section, steps S1-S6 are described in detail.

\noindent \textit{S1: Data Sets}\\
Two  separate sets of image data (i.e., two sets $\chi$) are used in this study, both brain MR images of patients with Multiple Sclerosis, one of them being a T2 weighted acquisition, and the other, a proton density (PD) weighted set, with the following acquisition parameters: Fast Spin Echo sequence, 1.5T GE Signa scanner, TR=2500 \textit{msec}, voxel size 0.86x0.86x3 $mm^3$. Each of the two sets is composed of 10 scenes. Since the two data sets for each patient are acquired in the same session with the same repetition time but by capturing different echos $(TE=18 msec, 96 msec)$, the T2 and PD scenes for each patient can be assumed to be in registration. 

\noindent \textit{S2. Non-uniformity Correction, Standardization}\\
For non-uniformity correction, we use the method based on the concept of local morphometric scale called g-\textit{scale}~(\cite{mada}). Built on fuzzy connectedness principles, the g-\textit{scale} at a voxel $\nu$ in a scene $\mathcal{C}$ is  the largest set of voxels fuzzily connected to $\nu$ in the scene such that all voxels in this set satisfy a predefined homogeneity criterion. Since the g-\textit{scale} set represents a partitioning of the scene domain $C$ into fuzzy connectedness regions by using a predefined homogeneity criterion, resultant g-\textit{scale} regions are locally homogeneous, and spatial contiguity of this local homogeneity is satisfied within the g-\textit{scale} region. 

g-\textit{scale} based non-uniformity correction is performed in a few steps as follows. First, g-\textit{scale} for all foreground voxels is computed. Second, by choosing the largest g-\textit{scale} region, background variation is estimated. Third, a correction is applied to the entire scene by fitting a second order polynomial to the estimated background variation. These three steps are repeated iteratively until the largest g-\textit{scale} region found is not significantly larger than the previous iteration's largest g-\textit{scale} region.

Standardization is a  pre-processing technique which maps non-linearly image intensity gray scale into a standard intensity gray scale through a training and a transformation step. In the training step, a set of images acquired for the same body region $B$ as per the same protocol $P$ are given as input to \textit{learn} histogram-specific parameters. In the transformation step, any given image for $B$ and $P$  is standardized with the estimated histogram-specific landmarks obtained from the training step. In the data sets considered for this study, $B=head$ and $P$ represents two different protocols, namely T2 and PD. The training and transformation steps are done separately for the two protocols.

The basic premise of standardization methods is that, in scenes acquired for a given $\left\langle B, P\right\rangle $, certain tissue-specific landmarks can be identified on the histogram of the scenes. Therefore, by matching the landmarks, one can standardize the gray scales. Median, mode, quartiles, and deciles, and intensity values representing the mean intensity in each of the largest few g-\textit{scale} regions have been used as landmarks. Additionally, to handle outliers, a ``low'' and ``high'' intensity value (selected typically at 0 and 99.8 percentiles) are also selected as landmarks.

In the training step, the landmarks are identified for each training scene specified for $\left\langle  B, P\right\rangle $ and intensities corresponding to the landmarks are mapped into an assumed standard scale. The mean values for these mapped landmark locations are computed. In the transformation step, the histogram of each given scene $\mathcal{C}$ to be standardized is computed, and intensities corresponding to the landmarks are determined. Sections of the intensity scale of $\mathcal{C}$ are mapped to the corresponding sections of the standard scale linearly so that corresponding landmarks of scene $\mathcal{C}$ match the mean landmarks determined in the training step. (The length of the standard scale is chosen in such a manner that the overall mapping is always one-to-one and no two intensities in $\mathcal{C}$ map into a single intensity on the standard scale.) Note that the overall mapping is generally not a simple linear scaling process but, indeed, a non-linear (piece-wise linear) operation; see~(\cite{std,std_new}) for details. In the present study, standardization is done separately for T2 and PD scenes. 

\begin{figure}[h]
 \begin{center}
   \begin{tabular}{c}
   \includegraphics[height=8cm]{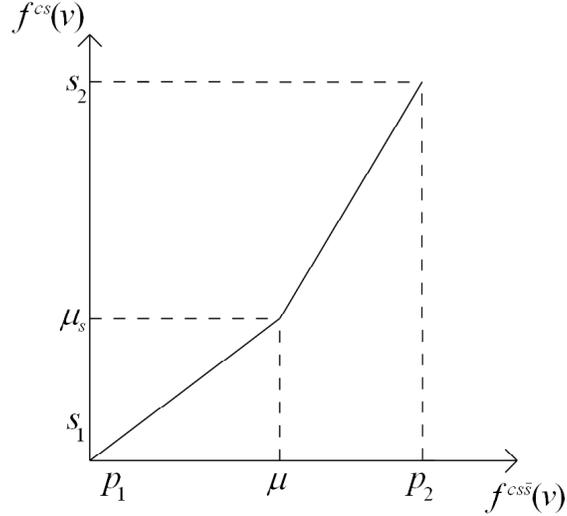} 
   \end{tabular}
   \end{center}
\caption{The standardization transformation function for inverse mapping with the various parameters shown.\label{img:mapping}}
\end{figure}

\noindent \textit{S3. Applying Non-standardness}\\
To artificially introduce non-standardness into a \textit{clean scene} $\mathcal{C}^{cs}=\left( C, f^{cs}\right) $, we use the idea of the inverse of the standardization mapping described in~(\cite{std_nu}). A typical standardization mapping is shown in Figure~\ref{img:mapping}. In this figure, only three landmarks are considered - ``low'' and ``high'' intensities $(p1$ and $p2)$ and the median $(\mu)$ corresponding to the foreground object. There are two linear mappings: the first from $[p_{1},\mu]$ to $[s_1,\mu_s]$ and the second from $[\mu, p_{2}]$ to $[\mu_s, s_2]$. $\left[ s_1, s_2\right] $ denotes the standard scale. The horizontal axis denotes the non-standard input scene intensity and vertical axis indicates the standardized output scene intensity. In inverse mapping, since $\mathcal{C}^{cs}$ has already been standardized, the vertical axis can be considered as the input scene intensity, $f^{cs}(\nu)$, and the horizontal axis can be considered as  the output scene intensity, $f^{cs\overline{s}}(\nu)$, where mapping the \textit{clean scene} through varying the slopes $m_1$ and $m_2$ results in non-standard scenes. By using the values of $m_1$ and $m_2$ 
within the range of variation observed in initial standardization mappings of corrected scenes, the non-standard scene intensities can be obtained by
 \begin{equation}
\label{eq:inversemapping}
f^{cs\overline{s}}(\nu)= \left\{ \begin{array}{ll}
\lceil \frac{f^{cs}(\nu)}{m_1} \rceil &
\textrm{if $f^{cs}(\nu) \leq \mu_s$}\\
\lceil  \frac{\left( f^{cs}(\nu)-\mu_s \right)}{m_2} + \mu  \rceil &
\textrm{if $f^{cs}(\nu) > \mu_s,$}
\end{array} \right.
\end{equation}
where  $\lceil . \rceil$ converts any number y$\in \Re$ to the closest integer Y, and $\mu_s$ denotes the median intensity on the standard scale. 

In order to keep the number of registration experiments manageable, this simple model was used which involves only two variables $m_1$ and $m_2$. Even so, as described later, this study entails nearly 20,000 registration experiments.
 
\noindent \textit{S4. Applying Affine Deformations}\\
All components of the affine transformation - rotations about all three axes and translation, scaling, and shear in all three directions - are taken into account in creating scene sets $\chi^{cst}$ and $\chi^{cs\overline{s}t}$.

\noindent \textit{S5. Scene Registration}\\
The algorithm that determines the affine transformation matrix by minimizing the sum of squared scene intensity differences as described in~(\cite{ash}) is used in this step. A separate transformation matrix is found for registering each $\mathcal{C}^{cs}$ to $\mathcal{C}^{cst}$, resulting in $T_s$, and $\mathcal{C}^{cs}$ to $\mathcal{C}^{cs\overline{s}t}$, resulting in $T_{ns}$.

\noindent \textit{S6. Evaluation}\\
Two types of tests were carried out - \textit{accuracy} and \textit{consistency}. The goal of the accuracy test was to determine how close the recovered registration transformations are to the known true transformations. The aim of the consistency test was to check how the observed accuracy behavior would consistently occur when different accuracy tests are conducted. In each test, two transformations  $T_s$ and $T_{ns}$ are compared by using the methodology that is described in~(\cite{reg_udupa}) and summarized below.  Let $\mathcal{C}_{T2}\in\chi_{T2}$ and $\mathcal{C}_{PD}\in\chi_{PD}$ be the T2 and PD scenes of any particular patient. Let $T_{s,x}$ and $T_{ns,x}$, $x\in\left\lbrace T2, PD\right\rbrace $, be the transformations obtained in Step S5 by matching $\mathcal{C}_x^{cs}$ to $\mathcal{C}_x^{cst}$ and $\mathcal{C}_x^{cs}$ to $\mathcal{C}_x^{cs\overline{s}t}$, respectively. In the \textit{accuracy} tests, $T_{s,x}$ and $T_{ns,x}$ are compared with the true (known) transformation $T$ over all levels of non-standardness and deformations that were employed. Both $T_{s,x}$ and $T_{ns,x}$ are expected to be the same as $T$. In the \textit{consistency} tests, $T_{s, T2}$ with $T_{s, PD}$ and $T_{ns, T2}$ with $T_{ns, PD}$ are compared over all levels of non-standardness and deformations that were applied, and they are expected to be equal because PD and T2 scenes of the same patient are already in registration as described in S1. We measure the error between two transformations (in both the above scenarios) by the root-mean-squared error (RMSE) for the eight corner voxels of the box that approximately bounds the head, i.e., the volume of interest in our application.

\begin{figure}[b]
 \begin{center}
   \begin{tabular}{c}
   \includegraphics[height=8cm]{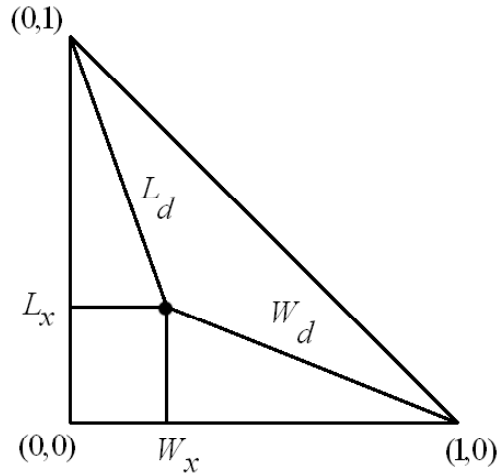} 
   \end{tabular}
   \end{center}
\caption{Mapping procedure for ``goodness'' value in normalized win-loss  $W_x-L_x$  plane. \label{img:merit}}
\end{figure}

In the accuracy test, for a given level of applied non-standardness and affine deformation, we get 20 pairs of RMSE values, each pair indicating how close $T_{s, x}$ and $T_{ns, x}$ are to the true transformation $T$. A paired t-test is conducted to compare the accuracy of the two transformations based on RMSE values. The outcome of this test will be that either of the two transformations is more accurate than the other or there is no significant difference between the two (throughout we use $P \leq 0.05$ to indicate statistical significance). The set of all levels of applied deformations is divided into three groups - small, medium, large. For each of 8 levels of applied non-standardness and under each  of these three groups, the number of occurrences of wins (w), losses (l) and non-significant differences (n) are counted for $T_{ns}$ over $T_s$. The number of wins and losses is normalized to get values in the range $[0, 1]$: $W_x=\frac{w}{w+l+n}$, $L_x=\frac{l}{w+l+n}$. A particular configuration of wins and losses can be identified by a point with coordinates $\left( W_x, L_x\right) $ in a win-loss triangle as in Figure~\ref{img:merit}. Large values of $L_d$ and small values of $W_d$ indicate that the performance point is closer to the point $\left( 1, 0\right) $ of all-wins. The following metric is used to express the ``goodness'' value of the configuration.
\begin{equation}
\gamma = \frac{L_d}{W_d}=\sqrt{\frac{\left( 1-L_x\right)^2 +W_x^2}{\left( 1-W_x\right)^2 +L_x^2}}.
\end{equation}

The procedure in the consistency test is similar to the above except that we have 10 pairs of RMSE values to compare and these values are obtained not by using the known true transformation but by using $T_{S,PD}$ for $T_{S,T2}$ and $T_{NS,PD}$ for $T_{NS,T2}$.

\section{Experimental Results}
\label{sec:results}

\subsection{Implementation Details}
\subsubsection{Correction and Standardization}
\begin{figure}[h]
 \begin{center}
   \begin{tabular}{c}
   \includegraphics[height=10 cm]{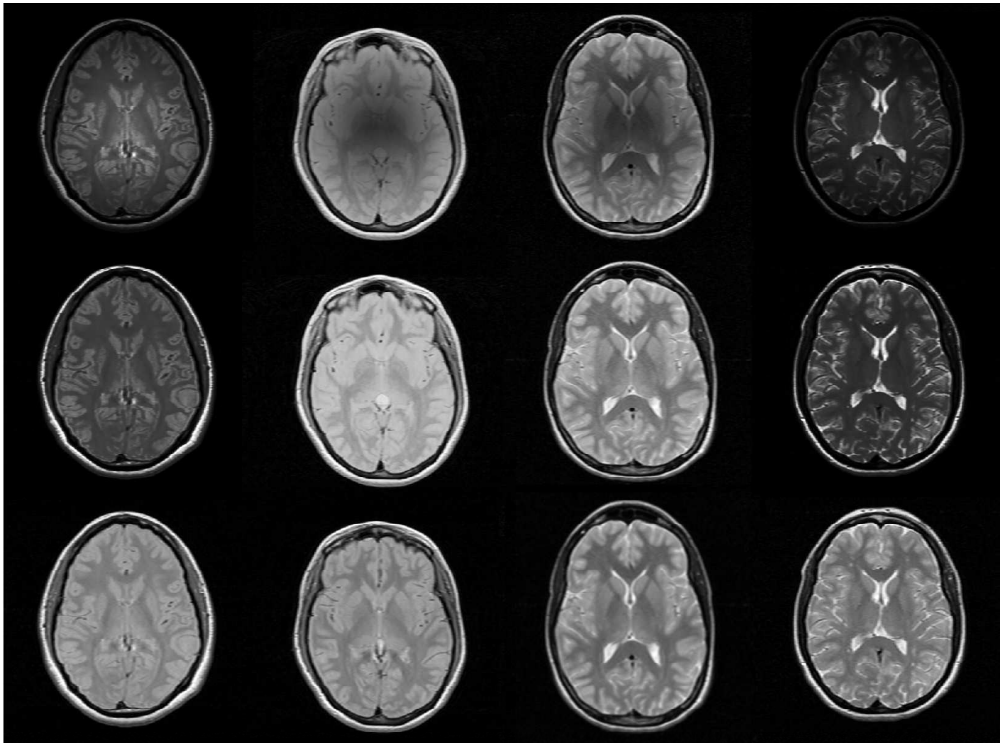} 
   \end{tabular}
   \end{center}
\caption{Two slices from PD (first and second columns) and two slices from T2-weighted scenes (third and fourth columns) selected from two different studies before correction and standardization are displayed at default windows in the first row. Corresponding slices of the g-scale corrected scenes are shown in the second row. Clean scenes obtained after the standardization process of the corresponding corrected scenes are displayed at a standard window in the third row. \label{img:cs}}
\end{figure}

These operations are carried out by using the 3DVIEWNIX software~(\cite{3dviewnix}). Based on the experiments in~(\cite{std, std_numeric, std_new}), minimum and maximum percentile values are set to $pc_{1}=0$ and $pc_2=99.8$, respectively. In the standard scale, $s_1$ and $s_2$ are set to $s_1=1$ and $s_2=4095$. Figure~\ref{img:cs} shows the original, corrected, and standardized (after correction) of two PD and two T2-weighted slices taken from two different studies in the first, second and third rows, respectively. The gain in the similarity of resulting image intensities for similar tissue types obtained can be readily seen.

\subsubsection{Adding known levels of non-standardness}
\begin{figure}[h]
 \begin{center}
   \begin{tabular}{c}
   \includegraphics[height=7 cm]{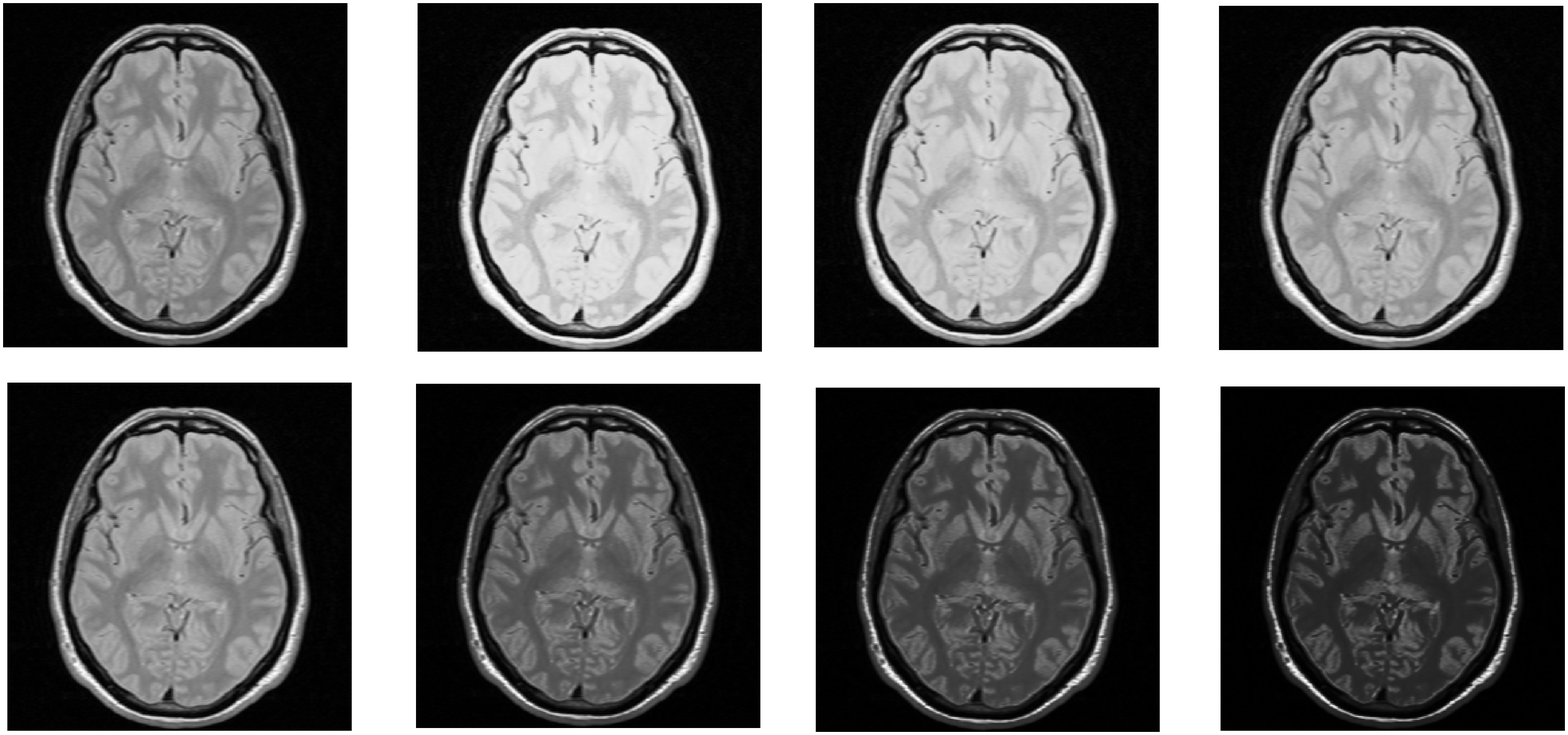} 
   \end{tabular}
   \end{center}
\caption{First image in the first row is a slice of a clinical PD weighted clean scene of the brain. The other slices are obtained by adding the 7 different levels of non-standardness into the clean scene (7 different levels of non-standardness are $\overline{\psi_1}$ to $\overline{\psi_7}$). All images are displayed at the fixed gray level window chosen for the clean scene. \label{img:pd}} 
\end{figure}

\begin{figure}[h]
 \begin{center}
   \begin{tabular}{c}
   \includegraphics[height=7 cm]{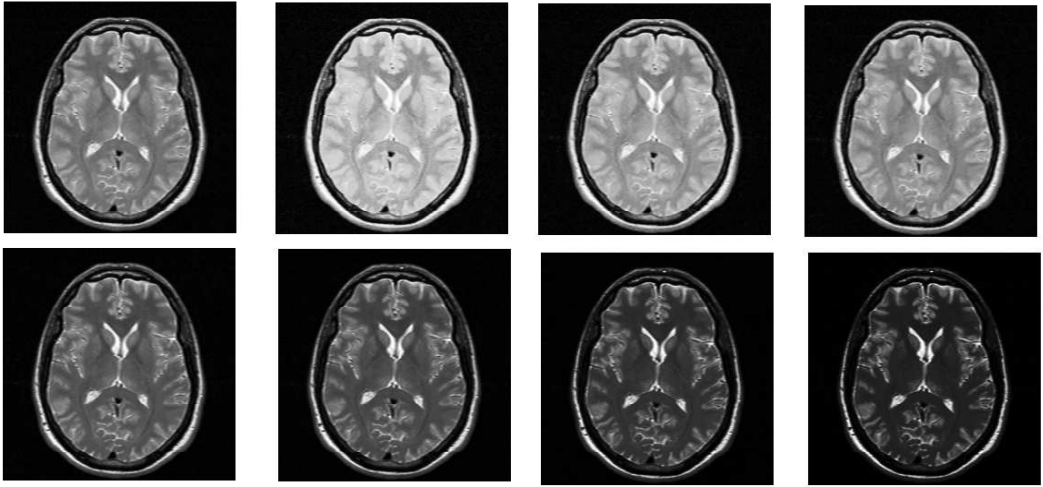} 
   \end{tabular}
   \end{center}
\caption{First image in the first row is a slice of a clinical T2 weighted clean scene of the brain. The other slices are obtained by adding the 7 different levels of non-standardness into the clean scene. (7 different levels of non-standardness are $\overline{\psi_1}$ to $\overline{\psi_7}$). All images are displayed at the fixed gray level window chosen for the clean scene. \label{img:t2}}
\end{figure}

We combine eight different ranges of the slopes $m_1$ and $m_2$ to introduce  small, medium, and large scale non-standardness into the scenes. This means that, for each \textit{clean scene}, we obtain eight scenes, one of which is the default \textit{clean scene} itself, two scenes consisting of small scale non-standardness, two scenes consisting of medium scale non-standardness, and three scenes consisting of large scale non-standardness. The ranges of applied non-standardness are given in Table~\ref{table:slopes}. We have arrived at these values by examining the training part of the standardization process through computing the ranges of the slopes $m_1$ and $m_2$ that are utilized in standardizing the corrected scenes. Figures~\ref{img:pd} and~\ref{img:t2} illustrate the process of introducing known levels of non-standardness into the \textit{clean} slices of a PD and a T2-weighted scene utilized in our study. In both figures, the first display shows the original clean slice and the rest show the resulting non-standard slices.

\begin{table}[h]
\caption{Description of the different range of the slopes $m_1,m_2$ for introducing artificial non-standardness\label{table:slopes}}
\begin{center}
\begin{tabular}{|c|c|c|}
\hline \hline
function & Range & Description \\ \hline \hline
$\overline{\psi}_1$   & $\left\lbrace 0.9 \leq m_1,m_2 \leq 1.5 \right\rbrace$  & \multirow{2}{*}{Small Scale} \\  
$\overline{\psi}_2$   & $\left\lbrace 0.6 \leq m_1,m_2 \leq 0.9 \right\rbrace$   &                      \\ \hline \hline
$\overline{\psi}_3$   & $\left\lbrace 1.5 \leq m_1,m_2 \leq 2.0 \right\rbrace$   & \multirow{2}{*}{Medium Scale}\\  
$\overline{\psi}_4$   & $\left\lbrace 2.0 \leq m_1,m_2 \leq 2.4 \right\rbrace$   &   \\ \hline \hline
$\overline{\psi}_5$   & $\left\lbrace 2.4 \leq m_1,m_2 \leq 2.7 \right\rbrace$   &   \multirow{3}{*}{Large Scale} \\ 
$\overline{\psi}_6$   & $\left\lbrace 2.7 \leq m_1,m_2 \leq 3,0 \right\rbrace$   &  \\ 
$\overline{\psi}_7$   & $\left\lbrace 3.0 \leq m_1,m_2 \leq 3.3 \right\rbrace$   &  \\ \hline \hline
\end{tabular}
\end{center}
\end{table}

\subsubsection{Applying known amounts of deformation}
Three different rotations (0, medium, and large angle), three translations (0, medium, and large displacement), three levels of scaling (0, medium, and large), and three levels of shearing (0, medium, and large) are combined to introduce 81 different known levels of deformation such that for all non-zero transformations, all three directions/axes are involved. Table~\ref{table:defval} summarizes the amount of these transformations  used for each axis in the  three different groups.

\begin{table}[h]
\caption{The amount of deformations corresponding to different groups of transformations. \label{table:defval}}
\begin{center}
\begin{tabular}{|c|c|c|c|}
\hline \hline
Transformation Type & Zero & Medium & Large \\ \hline \hline
Translation           &  $0$ pixels  &     $5$ pixels &  $20$ pixels \\ \hline
Rotation               &  $0^o$& $2^o$ & $6^o$\\   \hline
Scaling                 & $1$  & $1.05$ & $1.15$\\    \hline
Shearing               & $0$   & $0.01$   & $0.05$\\ \hline \hline
\end{tabular}
\end{center}
\end{table}

\subsubsection{Registration}
We use the scene based affine registration method made available in the SPM software~(\cite{ash}). The algorithm determines the affine transformation matrix  that optimally registers the two scenes by optimizing the sum of squared differences between scene intensities.

\subsection{Results}
For each of the 20 clean scenes in $\chi_{T2}^{cs} \bigcup \chi_{PD}^{cs}$, considering the 7 different levels of non-standardness together with one level of standardness (i.e. total of 8 levels) and 3 x 3 x 3 x 3 = 81 different levels of mis-registration, there will be 8 x 81 = 648 scenes. Thus, in the accuracy test, there will be 20 x 648 = 12,960 registration experiments. In the consistency test, similarly, there will be 6,480 additional registration experiments. These additional experiments  can be considered a validation for accuracy tests because they show how consistent the accuracy of the registration experiments are by using the fact that T2 and PD scenes are in registration.


The results of the comparison experiments are reported in Tables~\ref{table:goodness} and~\ref{table:goodness2} for accuracy and consistency tests, respectively,  for 7 sets of non-standard scenes with respect to the registration performance of \textit{clean scenes}. The tables summarize the effectiveness of the registrations for each type of deformation recovered. The goodness values indicate that the ability to recover known deformations from transformed scenes is lower if intensity variations between source and target scenes are large. The goodness value $\gamma < 1$ for scenes with non-standardness indicates that the registration between \textit{clean scenes} outperforms  registration between scenes with certain levels of non-standardness and  this is true only if $W_x<L_x$. 

\begin{table}[h]
\begin{center}
\caption{Comparison of methods for Accuracy. The Goodness values $\gamma$ are listed. Type of non-standardness are indicated by $\overline{\psi_1}, ..., \overline{\psi_7}$, and the type of affine deformations are indicated by small, medium, and large in the columns.  \label{table:goodness}}
\begin{tabular}{|c|c|c|c|c|}
\hline  Type of Non-Standardness  & Small  & Medium  & Large  & Total  \\ 
\hline  $\overline{\psi}_1 $ &  1 & 0.8222  &  0.6562  & 0.7811\\  
\hline  $\overline{\psi}_2 $ &  0.9400 &    0.8305 & 0.6167  &  0.7716\\ 
\hline  $\overline{\psi}_3 $ & 0.9369 &  0.7751 &  0.6309 & 0.7651 \\ 
\hline  $\overline{\psi}_4 $ & 0.9318 & 0.7004   & 0.6048  & 0.7565   \\ 
\hline  $\overline{\psi}_5 $  & 0.8806  & 0.6004  & 0.5622  & 0.6254  \\ 
\hline  $\overline{\psi}_6 $ & 0.7565 & 0.5511  &  0.5341  & 0.5881 \\ 
\hline  $\overline{\psi}_7 $ & 0.7447 &   0.5901  & 0.5051  & 0.5819 \\ \hline 
\end{tabular} 
\end{center}
\end{table}

\begin{table}[h]
\begin{center}
\caption{Comparison of methods for Consistency. The Goodness values $\gamma$ are listed. Type of non-standardness are indicated by $\overline{\psi_1}, ..., \overline{\psi_7}$, and the type of affine deformations are indicated by small, medium, and large in the columns. \label{table:goodness2}}
\begin{tabular}{|c|c|c|c|c|}
\hline  Type of Non-Standardness  & Small  & Medium  & Large  & Total  \\ 
\hline  $\overline{\psi}_1 $ &  1.3427 & 0.9289  &  0.7423  & 1\\  
\hline  $\overline{\psi}_2 $ &  1.1491 &    1 & 0.8039 &  0.9530\\ 
\hline  $\overline{\psi}_3 $ & 1 &  0.8039 & 0.7423 & 0.8417 \\ 
\hline  $\overline{\psi}_4 $ & 0.8622 & 0.7447   & 0.5434  & 0.7062   \\ 
\hline  $\overline{\psi}_5 $  &   0.9289 &  0.6722 &  0.5023 &  0.6934 \\ 
\hline  $\overline{\psi}_6 $ & 0.8636  &  0.6890  & 0.5023    & 0.6722  \\ 
\hline  $\overline{\psi}_7 $ & 1.0752 &   0.7097  & 0.2998  & 0.6468 \\ \hline 
\end{tabular} 
\end{center}
\end{table}

A strong possible reason for the better performance of \textit{clean scenes} with respect to the non-standard scenes is that structural information for the same subject in different non-standardness levels is not  the same. Therefore, correlations of the intensity values for each structure in the scenes may not reach the optimum to which the registration algorithm converges. Registration parameters are obtained through maximizing the similarity of two scenes, and it is well demonstrated  in Tables~\ref{table:goodness} and~\ref{table:goodness2}  that correlation of the intensities is maximum when each structure in the scene has fixed tissue specific meaning. 

\begin{figure}[h]
 \begin{center}
   \begin{tabular}{c}
   \includegraphics[width=12 cm]{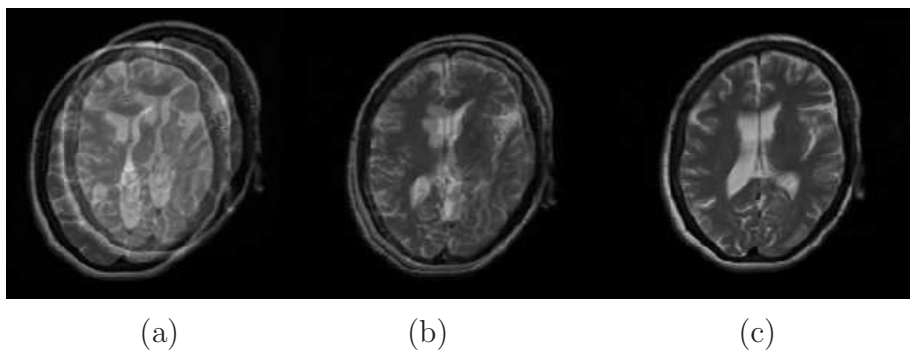} 
   \end{tabular}
   \end{center}
(a) \quad \quad \quad \quad \quad \quad \quad  (b)  \quad \quad \quad \quad \quad \quad \quad \quad  \quad (c)
\caption{(a) An example of registration failure for \textit{clean scene} with a large deformation. (b) An example for registration success for non-standard scene for the same amount of deformation as in the example in (a). (c) If a proper initialization matrix is given as input to the registration algorithm, clean scene registration performance becomes better than the example for non-standard scene. \label{img:uclu}}
\end{figure}

Another possible reason is that the relationship between voxel intensities may be non-linear. Since the introduction of non-standardness is itself a non-linear process, the similarity function is likely to be affected by this situation in the form of local fluctuations which may even lead to not only less accurate registration results but also to the situation of the optimization process getting locked at local optima. The opposite situation may happen as well, especially for large scale deformations; the registration algorithm may easily fail regardless of the standardization level of the scenes (see Figure~\ref{img:uclu} (a) for a failing example of \textit{clean scenes}). Local fluctuations in the similarity measure due to non-standardness may lead to different optimum points depending on the degree of non-standardness, some of which may even improve registration, as shown in Figure~\ref{img:uclu} (b). Although the registration algorithm did not get stuck in the latter case, the accuracy of the registration quality was not high especially in terms of translation parameters. In order to cope with possible failing examples in registration, we ran the registration algorithm with proper initial estimation of the transformation matrix rather than the default identity transformation matrix used in all experiments. Figure~\ref{img:uclu} (c) shows the registered and transformed \textit{clean scenes} overlaid where fuzziness in gray scale images demonstrates the misalignment. Compared to the registration of non-standard scene in Figure~\ref{img:uclu} (a), the performance of the registration of \textit{clean scene} in this example is still better when the initial estimation of transformation matrix is given as input to the registration process.

Based on the fact that similarity of a pair of registered \textit{clean scenes} is higher than the similarity of a pair of  registered non-standard scenes,  it can be deduced that substantially improved uniformity of tissue meaning between two scenes of the same subject being registered improves registration accuracy. Our experimental results demonstrate that scenes are registered better whenever the same tissues are represented by the same intensity levels. 

We note that, in both tables, most of the entries are less than 1. This indicates that in both accuracy and consistency tests, the standardized scene registration task wins over the registration of non-standard scenes. Table~\ref{table:goodness}  on its own does not convey any information about what the actual accuracies in the winning cases are, or about whether the win happens for T2 scenes only, PD only, or for both. The fact that a majority of the corresponding cells in these tables both indicate wins suggests that accuracy-based wins happen for both T2 and PD scenes. Conversely, a favorable $\gamma$ value in Table~\ref{table:goodness2} does not convey any information about whether the high consistency indicated also signals accuracy. Thus, accuracy and consistency are to some extent independent factors, and they together give us a more complete picture of the influence of non-standardness on registration.

\section{Concluding Remarks}
\label{sec:conc}
We described a controlled environment for determining the effects of intensity standardization on registration tasks in which the best image quality (\textit{clean scene}) was established by the sequence of correction operation followed by standardization. We introduced several different levels of non-standardness into the \textit{clean scenes} and performed nearly 20,000 registration experiments for small, medium and large scale deformations. We compared the registration performance of \textit{clean scenes} with the performance of scenes with non-standardness and summarized the resulting goodness values. From overall accuracy and consistency test results in Tables~\ref{table:goodness} and~\ref{table:goodness2}, we conclude that intensity variation between scenes degrades registration performance. Having tissue specific numeric meaning in intensities maximizes the similarity of images which is the essence of the optimization procedure in registration. Standardization is therefore strongly recommended in the registration of images of patients in any longitudinal and follow up study, especially when image data come from different sites and different scanners of the same or different brands.

In this paper, we introductorily addressed the problem of the potential influence of intensity non-standardness on registration. This is indeed a small segment of the much larger full problem: Unlike the specific intra-modality (or intra protocol) registration task considered here, there are many situations in which the source and the target images may be from different modalities or protocols (e.g., CT to MRI, PET to MRI, and T1 to T2 registration etc.), and each such situation may have its own theme of non-standardness. Further, these themes may depend on the body region, the scanner, and its brand. We determined that a full consideration of these aspects was just beyond the scope of this paper. Since the sum of squared differences is one of the most appropriate similarity metrics for intra-modality registration, we focused on this metric in our study. But, clearly, more studies of this type in the more general settings mentioned above are needed.


Thus far, we controlled the computational environment via two factors: standardization and correction. A third important factor, noise, can be also embedded into the framework. It is known that correction itself introduces non-standardness into the scenes and it also enhances noise. Investigating the interrelationship between correction and noise suppression algorithms and determining the proper order for these operations has been studied recently~(\cite{montillo}). A question immediately arises as to how standardization affects registration accuracy for different orders of correction and noise filtering.  Based on the study in~(\cite{montillo}), we may conclude that non-uniformity correction should precede noise suppression and that standardization should be the last operation among the three to obtain best image quality. However, it remains unclear as to how a combination of deterministic methods (standardization and correction) affects a random phenomenon like noise. It is thus important to study these three phenomena in the future on their own or in relation to how they may influence the registration process, especially in multi-center studies wherein data come from different scanners and brands of scanners.

\section{Acknowledgement}
This paper is presented in SPIE  Medical Imaging - 2010. The complete version of this paper is published in Elsevier Pattern Recognition Letters, Vol(31), pp.315--323, 2010.

This research is partly funded by the European Commission Fp6 Marie Curie Action Programme (MEST-CT-2005-021170). The second author's research is funded by an NIH grant EB004395.



\end{document}